\documentclass[runningheads]{llncs}
\usepackage{amsmath}
\usepackage{graphicx}
\usepackage{xcolor}
\usepackage{hyperref}
\usepackage{subfig}
\begin{document}
\title{Inferring Causal Direction from Observational Data: A Complexity Approach}
\titlerunning{Inferring Causal Direction from Observational Data}
%
\author{Nikolaos  Nikolaou\inst{1}\orcidID{0000-0001-8453-7574} \and\\
Konstantinos Sechidis\inst{2}
}
\authorrunning{N. Nikolaou \& K. Sechidis}
%
\institute{Department of Physics and Astronomy, University College London\\ Gower Street, London, WC1E 6BT, UK\\
\email{n.nikolaou@ucl.ac.uk}
\and
Novartis International AG\\
\email{AAA@novartis.com}
}
\maketitle              
\begin{abstract}
At the heart of causal structure learning from observational data lies a deceivingly simple question: given two statistically dependent random variables, which one has a causal effect on the other? This is impossible to answer using statistical dependence testing alone and requires that we make additional assumptions. We propose several fast and simple criteria for distinguishing cause and effect in pairs of discrete or continuous random variables. The intuition behind them is that predicting the effect variable using the cause variable should be `simpler' than the reverse -- different notions of `simplicity' giving rise to different criteria. We demonstrate the accuracy of the criteria on synthetic data generated under a broad family of causal mechanisms and types of noise.

\end{abstract}

\section{Introduction}
Recent advancements in machine learning have enabled the efficient training of powerful statistical models from large amounts of high-dimensional data. Yet, despite their many successes, learning systems are still almost exclusively operating on the level of \emph{statistical associations} among the observed variables~\cite{pearl2009causality}.

The next big step in the field will involve \emph{causal inference}; moving beyond simply capturing statistical associations to modelling \emph{cause} and \emph{effect} relationships among the underlying variables. Fields like healthcare \& epidemiology~\cite{glass2013causal,vandenbroucke2016causality}, bioinformatics \& pharmaceutical research~\cite{li2013pathway,osimani2015causal,yazdani2015causal,triantafillou2017predicting,badsha2019learning}, policy-making in social sciences~\cite{marini1988causality,sobel2000causal}, energy \& climate~\cite{ferkingstad2011causal,hannart2018probabilities}, economics \& finance~\cite{hoover2012economic,varian2016causal} and the physical sciences~\cite{wood2015lesson,foreman2015systematic} will be among the first to benefit from the coming advancements in \emph{causal modelling}. 

The first challenge to reasoning about cause and effect lies in the construction of a causal model of the variables involved in a given problem. Learning the underlying causal structure from observational data --i.e. without performing a randomized experiment-- is a challenging problem  as it practically entails leaping from correlation to causation.

This work demonstrates that such a leap can be justified -- under some assumptions-- by using a simple intuition: predicting the effect variable using the cause variable (i.e. respecting the true causal direction) should be `\emph{simpler}' than the inverse. Different definitions of `simplicity' yield different criteria for distinguishing cause from effect. We present a number of such criteria all of which are easy to implement, fast and applicable to both discrete \& continuous random variables (r.v.'s).
We demonstrate their accuracy on synthetic data generated under a broad family of causal mechanisms and types of noise.

\section{Background}
\subsection{Causal Structure Learning \& the 2-variable Problem}

Let $X_i$, $i=1, \dots, n$ be r.v.'s, the causal relationships among which are captured by a \emph{Structural Causal Model (SCM)} \cite{spirtes2000causation,pearl2009causality} with an underlying \emph{causal graph} $\mathcal{G}$, a Directed Acyclic Graph, the vertices of which correspond to the random variables $X_i$ and an edge $X_i \rightarrow X_j$ denotes that $X_i$ has a direct causal effect on $X_j$.

The SCM models each $X_i$ as the result of an assignment, 
\begin{equation}
 X_i:= f_i(PA_i, U_i),~~~~i=1, \dots, n,
\end{equation}
where 
$f_i$ is a deterministic function  of $X_i$’s parents in $\mathcal{G}$ (denoted by $PA_i$) and $U_i$ a stochastic unexplained variable (i.e. a noise variable). The set of noises $U_i$, $i=1, \dots, n$ are assumed to be jointly independent\footnote{If $\exists$ $U_i \not \!\perp\!\!\!\perp U_j$, $i \neq j$, then by the \emph{Common Cause Principle} (see below), there exists another variable that explains their dependence and $\mathcal{G}$ is thus not causally sufficient.}.

Causal structure learning, --i.e. inferring the underlying graph $\mathcal{G}$-- from observational data is a non-trivial problem. Observational data can inform us of statistical (in)dependences among variables but, this alone is not sufficient for uncovering the underlying causal graph. The reason is that a single independence relationship involving a set of variables can correspond to several underlying subgraph structures involving them.

The most characteristic example is determining the causal direction in the case of two random variables $X$, $Y$ that are statistically dependent. The \emph{Common Cause Principle}~\cite{reichenbach1991direction} captures the connection between causality and statistical dependence. It states that if two observables $X$ and $Y$ are statistically dependent, then there exists a variable $Z$ that causally influences both and explains all the dependence in the sense of making them independent when conditioned on $Z$. As a special case, this variable can coincide with $X$ or $Y$. In other words, the causal structures $X \rightarrow Y$, $Y \rightarrow X$ and $X \leftarrow Z \rightarrow Y$ are all consistent with observing the statistical dependence $X \not \!\perp\!\!\!\perp Y$.

\subsection{Causal Direction, Complexity \& Minimum Description Length}

Does this mean that inferring causal direction from observational data is impossible? Not if we make additional assumptions. More concretely, let us assume we observe a statistical dependence $X \not \!\perp\!\!\!\perp Y$ and want to determine whether the causal direction is $X \rightarrow Y$ or $Y \rightarrow X$\footnote{We shall exclude from consideration the case of their dependence is a result of pure chance or that it is due to a third variable $Z$ s. t. $X \leftarrow Z \rightarrow Y$.}. It would be reasonable\footnote{Philosophically, the intuition that the true causal mechanism is the simplest among all possible ones can be justified by the \emph{Occam's Razor Principle}.} to assume that \emph{the true causal direction is the one that is the least complex to model}.

\emph{Complexity} can be interpreted in several ways, but ultimately the central argument is a general application of the \emph{Minimum Description Length (MDL) Principle}\cite{grunwald1998} which views learning as data compression: Learning to predict effect from cause
should be simpler than the other way round amounts to the dataset being easier to compress when modelled in the causal direction~\cite{janzing2012information,scholkopf2019causality}.

Consider the (lossless) two-stage code that encodes a variable $Y$ with \emph{length} $L(Y)$ by first encoding a hypothesis $f_{X \rightarrow Y}$ (modelling $Y$ as a function of $X$) in the set of considered hypotheses ${\cal{F}}$ and then coding $Y$ ``with the help of" $f_{X \rightarrow Y}$; in the simplest context this just means ``encoding the deviations of $Y$ in the data from the predictions made by $f_{X \rightarrow Y}$". The description length of $Y$ is then:
\begin{equation}
L(Y)=\min_{f_{X \rightarrow Y}\in{\cal {F}}} \lbrace L(f_{X \rightarrow Y})+L(Y|f_{X \rightarrow Y}) \rbrace.
\end{equation}
Similarly, encoding $X$ with the aid of $Y$ (i.e. ``encoding the deviations of $X$ in the data from the predictions made by $f_{Y \rightarrow X}$) we get:
\begin{equation}
L(X)=\min_{f_{Y \rightarrow X}\in{\cal {F}}} \lbrace L(f_{Y \rightarrow X})+L(X|f_{Y \rightarrow X}) \rbrace.
\end{equation}
Specifying a  ``language" for measuring the description length $L$ in practice (this is where our notion of complexity comes into play), we can reason that the true causal direction is $Y \rightarrow X$, if $L(Y) < L(X)$ or $X \rightarrow Y$, if $L(Y) > L(X)$. If $L(Y) = L(X)$, we do not have enough information to decide~\cite{vreeken2015causal,budhathoki2018origo}.

\section{Simple Criteria for Distinguishing Cause \& Effect}

Using the framework described above, we will now present simple criteria for determining causal direction between two r.v.'s $X$ and $Y$. The goal is to compare the complexity of a model that uses as input $X$ to predict $Y$, denoted as $f_{X \rightarrow Y}$ against that of model that uses as input $Y$ to predict $X$, denoted as $f_{Y \rightarrow X}$. 

For such a comparison to be meaningful, both models need to be drawn from the same model family $\cal{F}$, e.g. by being generated by the same learning algorithm and hyperparameter configuration. For the purposes of this work, we opted for using \emph{Decision Trees of unbounded depth} due to their ease of implementation and plethora of ways of characterizing their complexity. The criteria $J_{RE}$ \& $J_{IH}$ are not limited to this specific modelling choice. The rest, are measures of complexity specific to decision trees and must be appropriately replaced should $f_{X \rightarrow Y}$ \& $f_{Y \rightarrow X}$ be drawn from an alternative model family\footnote{E.g. if the models are obtained via polynomial regression, the degree of the resulting polynomial or the number of its coefficients are naive measures of complexity.}.

Continuous data must be discretized so that both $X$ \& $Y$ have the same cardinality (i.e. if both are continuous, we can use equal width binning.


For any criterion $J$, the rule for deciding causal directionality is the same:
\begin{equation}
\label{eq:J_decision}
\begin{split}
If~~~J~~~
\begin{cases}
>0 &\text{,~~~then predict $X \rightarrow Y$}\\
<0 &\text{,~~~then predict $Y \rightarrow X$}\\
=0 &\text{,~~~then abstain from prediction}
\end{cases}
\end{split}
\end{equation}
\subsubsection{Model Complexity: Tree Depth} The most straightforward way to compare the complexity of trees is comparing their depths. The intuition is that the tree of smallest depth is the simplest, hence the one capturing the true causal direction.
\begin{equation}
\label{eq:J_TD}
    J_{TD} = Depth(f_{X \rightarrow Y}) - Depth(f_{Y \rightarrow X})
\end{equation}
\subsubsection{Model Complexity: Tree Nodes} Following the same intuition, we can instead opt to use the total number of nodes as a proxy of complexity. Note that the cardinalities of $X$ \& $Y$ will affect the total number of nodes of $f_{X \rightarrow Y}$ and $f_{Y \rightarrow X}$. 
\begin{equation}
\label{eq:J_TN}
    J_{TN} = Number~of~Nodes(f_{X \rightarrow Y}) - Number~of~Nodes(f_{Y \rightarrow X})
\end{equation}
\subsubsection{Model Complexity: Tree Leaves} Along the same reasoning,  another option is to use the total number of leaf nodes to measure tree complexity. 
\begin{equation}
\label{eq:J_TL}
    J_{TL} = Number~of~Leaves(f_{X \rightarrow Y}) - Number~of~Leaves(f_{Y \rightarrow X})
\end{equation}
\subsubsection{Model Complexity: Path Length} The final tree-specific measure of complexity we examined is the mean path length. This is the average number of nodes a datapoint in the sample traverses before it reaches a leaf node\footnote{For comparison, the depth of the tree corresponds to the maximal path length.}.   
\begin{equation}
\label{eq:J_PL}
    J_{PL} = Mean~Path~Length(f_{X \rightarrow Y}) - Mean~Path~Length(f_{Y \rightarrow X})
\end{equation}
\subsubsection{Residual Entropy}
The intuition for this criterion is that if $X \rightarrow Y$ then the decrease in Shannon entropy $H$ afforded by modelling $Y$ as a function of $X$, $f_{X \rightarrow Y}(X)$ must be higher than the decrease in Shannon entropy afforded by modelling $X$ as a function of $Y$, $f_{Y \rightarrow X}(Y)$.
\begin{equation}
\label{eq:J_RE}
    J_{RE} = [H(Y) - H(Y - f_{X \rightarrow Y}(X)] - [H(X) - H(X - f_{Y \rightarrow X}(Y))]
\end{equation}
\subsubsection{Interpolation Hardness}
Another approach is to directly compare the ability of $f_{X \rightarrow Y}$ and $f_{Y \rightarrow X}$ to fit the training data, i.e. to be used to predict the target.
\begin{equation}
\label{eq:J_IH}
    J_{IH} = Loss(Y, f_{X \rightarrow Y}(X)) - Loss(X, f_{Y \rightarrow X}(Y)),
\end{equation}
where $Loss$ denotes some appropriate loss function, e.g. for discrete r.v.'s, the misclassification error or for continuous r.v.'s the Mean Squared Error (MSE).

\section{Empirical Evaluation}

We generate 1000 datasets, each consisting of 1000 $(X,Y)$ pairs under the SCM:
\begin{align*} 
X &=  U_X \\ 
Y &= f_Y(X, U_Y),
\end{align*}
where $U_X$ and $U_Y$ are independent noise variables that can be either:
\begin{itemize}
\item [1.] Discrete (resulting in discrete $X$ \& $Y$), in which case either can be:
\begin{itemize}
\item a. Uniform, drawn from $U(-R/2, R/2 + 1)$, $R$ being the r.v.'s cardinality.
\item b. Discretized Gaussian, drawn from $\mathcal{N}(0,1)$, then  discretized to $R$ bins.
\end{itemize}
\item [2.] Continuous (resulting in continuous $X$ \& $Y$), in which case either can be:
\begin{itemize}
\item a. Uniform, drawn from $U(-1, 1)$.
\item b. Gaussian, drawn from $\mathcal{N}(0,1)$.
\end{itemize}
\end{itemize}

And $f_Y$ is a deterministic function that can be $f_Y(X, U_Y) = f_Y^{X}(X) + f_Y^{N}(U_Y)$ or $f_Y(X, U_Y) = f_Y^{X}(X) * f_Y^{N}(U_Y)$ (additive or multiplicative noise), with both $f_Y^{X}$ and $f_Y^{N}$
being polynomials of degree $\in [1, 5]$ whose coefficients are integers $\in [-10, 10]$. Both the degree and the coefficients are drawn uniformly from their respective sets. The causal direction is flipped with probability 0.5. 

\begin{figure}[h]%
    \centering
    \subfloat[Tree Depth ($J_{TD}$)]{{\includegraphics[scale=.33]{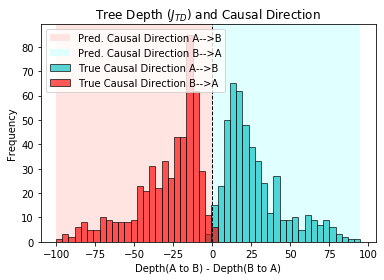} }}%
    \subfloat[Path Length ($J_{PL}$)]{{\includegraphics[scale=.33]{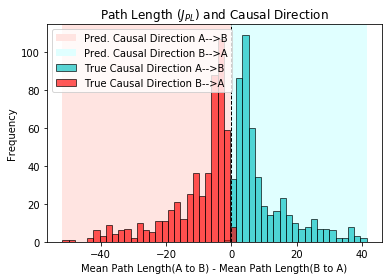} }}%
    \subfloat[Tree Nodes ($J_{TN}$)]{{\includegraphics[scale=.33]{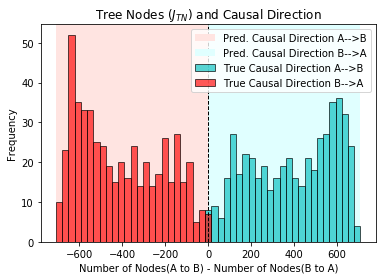} }}%
    \hfill
    \subfloat[Tree Leaves ($J_{TL}$)]{{\includegraphics[scale=0.33]{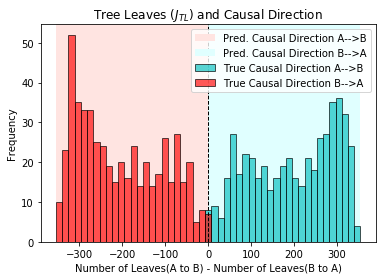} }}%
    \subfloat[Residual Entropy ($J_{RE}$)]{{\includegraphics[scale=0.33]{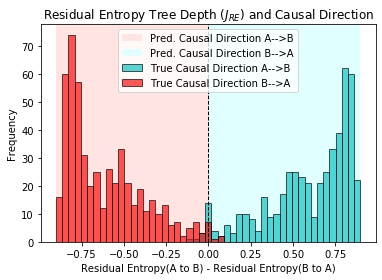} }}%
    \subfloat[Interpolation Hardness  ($J_{IH}$) ]{{\includegraphics[scale=0.33]{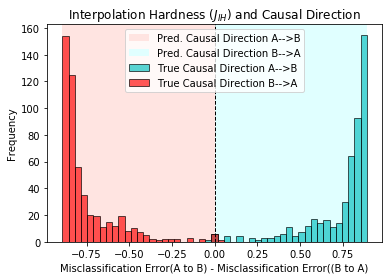} }}%
    \caption{Scores of each criterion on discrete r.v.'s, under additive noise. Noise variables follow uniform distributions with $R=20$. Histogram is color coded by true causal direction. Shaded regions show predicted causal direction under the criterion, dashed line its threshold. $J_{RE}$ is normalized by target r.v.'s entropy. Scores binned to $50$ equal width bins.}%
    \label{fig:disc}%
\end{figure}

\begin{figure}[h]%
    \centering
    \subfloat[Tree Depth ($J_{TD}$)]{{\includegraphics[scale=.33]{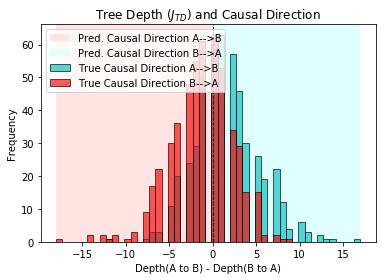} }}%
    \subfloat[Path Length ($J_{PL}$)]{{\includegraphics[scale=.33]{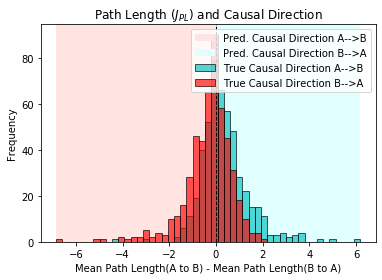} }}%
    \subfloat[Tree Nodes ($J_{TN}$)]{{\includegraphics[scale=.33]{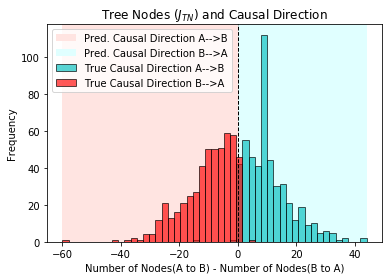} }}%
    \hfill
    \subfloat[Tree Leaves ($J_{TL}$)]{{\includegraphics[scale=0.33]{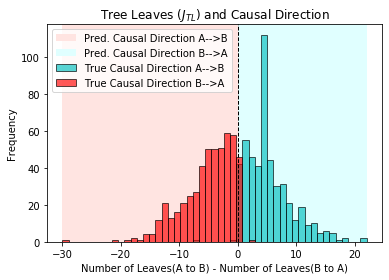} }}%
    \subfloat[Residual Entropy ($J_{RE}$)]{{\includegraphics[scale=0.33]{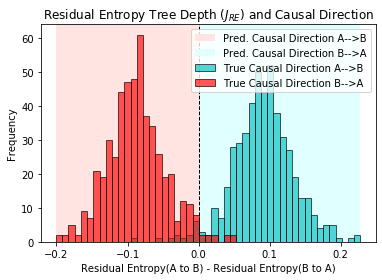} }}%
    \subfloat[Interpolation Hardness  ($J_{IH}$) ]{{\includegraphics[scale=0.33]{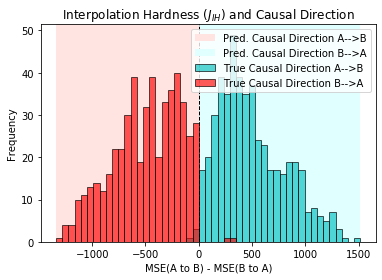} }}%
    \caption{Scores of each criterion on continuous r.v.'s under additive noise; noise variables follow uniform distributions -- see Fig. (1) for more details.}.%
    \label{fig:cont}%
\end{figure}

For each dataset, we calculate the value of each of the proposed criteria and predict the causal direction based on it\footnote{All code can be found at: \url{https://github.com/nnikolaou/CausalDirectionality}.}. Continuous r.v.'s are discretized to $100$ equal width bins. Tables \ref{tab:disc} \& \ref{tab:cont} show the mean accuracy of each criterion across datasets, its accuracy ignoring cases in which it abstains, the mean value of the criterion for the causal and for the anti-causal direction. Figures \ref{fig:disc} \& \ref{fig:cont} show the distribution of the values of each criterion for the causal and anti-causal direction. The dashed black line demarcates the decision threshold. Due to space limitations, we only present some of the results.

On discrete pairs of $X$ \& $Y$, all criteria are very accurate, regardless of the distributions of the noise variables.
Increasing the entropy of the cause r.v. (e.g. by increasing its cardinality $R$), only slightly decreases accuracy, but not considerably.
Increasing the entropy of the effect r.v. (e.g. by increasing the cardinality $R$ of its underlying noise variable) increases accuracy to 1.
For, continuous $X$, $Y$, criteria based on tree depth ($J_{TD}$, $J_{PL}$)
behave poorly (still above chance at predicting causal direction).
Criteria based on tree width ($J_{TN}$, $J_{TL}$) must be flipped - i.e. modelling effect using cause is associated with larger tree width; Once this is done, they perform as accurately as in the discrete case.
Criteria based on residuals ($J_{RE}$, $J_{IH}$) are very accurate. The above hold regardless of the noise distribution. Interestingly, when the noise is multiplicative the accuracy is higher -- almost perfect for all criteria.

\begin{table}[]
\center
\begin{tabular}{c||c|c|c|c|c|c}
\multicolumn{1}{l}{} & \multicolumn{6}{c}{Criterion} \\
\multicolumn{1}{l}{} & $J_{TD}$ & $J_{TN}$ & $J_{TL}$ & $J_{PL}$ & $J_{RE}$ & $J_{IH}$ \\ \hline \hline
Accuracy & 0.988 & 0.986 & 0.986 & 0.989 & 0.974 & 0.986\\ \hline
\begin{tabular}[c]{@{}c@{}}Accuracy\\ when not\\ Abstaining\end{tabular} & 0.995 & 0.998 & 0.998 & 0.996 & 0.986 & 0.998\\ \hline
\begin{tabular}[c]{@{}c@{}}Average value\\ Causal\end{tabular} & 9.252 & 39.000 & 20.000 & 6.883 & 0.214 & 0.865\\ \hline
\begin{tabular}[c]{@{}c@{}}Average value\\ Anti-causal\end{tabular} & 37.417 & 437.215 & 219.107 & 16.864 & 0.819 & 0.118        
\end{tabular}
\caption{Results of criteria on discrete $X$ \& $Y$, under additive noise. Noise r.v.'s are both uniform with $R=20$. $J_{RE}$ is normalized by target r.v.'s entropy.}
\label{tab:disc}
\end{table}

\begin{table}[]
\center
\begin{tabular}{c||c|c|c|c|c|c}
\multicolumn{1}{l}{} & \multicolumn{6}{c}{Criterion} \\
\multicolumn{1}{l}{} & $J_{TD}$ & $J_{TN}$ & $J_{TL}$ & $J_{PL}$ & $J_{RE}$ & $J_{IH}$ \\ \hline \hline
Accuracy & 0.583 & 0.909 & 0.909 & 0.628 & 0.976 & 0.990\\ \hline
\begin{tabular}[c]{@{}c@{}}Accuracy\\ when not\\ Abstaining\end{tabular} & 0.665 & 0.998 & 0.997 & 0.631 & 0.978 & 0.997\\ \hline
\begin{tabular}[c]{@{}c@{}}Average value\\ Causal\end{tabular} & 12.871 & 198.766 & 99.883 & 8.905 & 0.024 & 920.909\\ \hline
\begin{tabular}[c]{@{}c@{}}Average value\\ Anti-causal\end{tabular} & 14.314 & 188.454 & 94.727 & 9.251 & 0.113 & 427.146        
\end{tabular}
\caption{Results of criteria on continuous $X$ \& $Y$, under additive noise. Noise r.v.'s are both uniform. $J_{RE}$ is normalized by target r.v.'s entropy.}
\label{tab:cont}
\end{table}

\section{Conclusion \& Future Work}

We demonstrated that inferring causal direction from observational data is possible, if we make the --justified by Occam's Razor-- assumption that predicting the effect using the cause should be simpler than the other way round. 

We used decision trees to compare the difficulty of modelling $X$ using $Y$ vs. $Y$ using $X$. The resulting criteria we proposed address simplicity via the trees' structure, the entropy of their outputs or the quality of their fit. They are simple to implement, fast to compute and capable of handling both discrete and continuous variables. They were found to be highly accurate on a broad class of underlying causal mechanisms and noise types.  

The results suggest that there are important differences between discrete and continuous features. For instance the modelling direction producing the tree of larger width tends to coincide with the true causal direction in the case of continuous variables and the anti-causal one for discrete r.v.'s. This suggests that width --in the case of continuous r.v.'s-- captures aspects of fitting, not of redundant complexity and will be explored further. 

A more detailed theoretical analysis and unified treatment of the criteria presented in this paper is left for an extended version of this work. So is its application to scenarios involving more than two variables, mixed (discrete \& continuous) r.v.'s, a richer set of underlying causal mechanisms and applications to real world data.


%
%
%
%
\bibliographystyle{splncs04}
\bibliography{mybibliography}

\end{document}